\documentclass{article}

\usepackage{arxiv}

\usepackage[utf8]{inputenc} 
\usepackage[T1]{fontenc}    
\usepackage{hyperref}       
\usepackage{url}            
\usepackage{booktabs}       
\usepackage{amsfonts}       
\usepackage{nicefrac}       
\usepackage{microtype}      
\usepackage{cleveref}       
\usepackage{lipsum}         
\usepackage{graphicx}
\usepackage{natbib}
\usepackage{doi}

\title{Embedding-based Retrieval with LLM for Effective Agriculture
Information Extracting from Unstructured Data}

\date{}

\newif\ifuniqueAffiliation
\uniqueAffiliationtrue

\ifuniqueAffiliation 
\author{Ruoling Peng \\
	Department of Computer Science\\
	University of Sheffield\\
	Sheffield, Uk\\
	\And
	Kang Liu \\
	Department of Computer Science\\
	University of Sheffield\\
	Sheffield, Uk\\
	\And
	Po Yang \\
	Department of Computer Science\\
	University of Sheffield\\
	Sheffield, Uk\\
	\And
	Zhipeng Yuan \\
	Department of Computer Science\\
	University of Sheffield\\
	Sheffield, Uk\\
	\And
	Shunbao Li \\
	Department of Computer Science\\
	University of Sheffield\\
	Sheffield, Uk\\
}
\else
\usepackage{authblk}

\newbox{\orcid}\sbox{\orcid}{\includegraphics[scale=0.06]{orcid.pdf}} 
\fi
\renewcommand{\shorttitle}{\textit{arXiv} preprint}

\hypersetup{
pdftitle={Embedding-based Retrieval with LLM for Effective Agriculture Information Extracting from Unstructured Data},
pdfkeywords={Agriculture \and Embedding-based Retrieval \and Large Language Model \and Information Extraction},
}

\begin{document}
\maketitle

\begin{abstract}
	Pest identification is a crucial aspect of pest control in agriculture.
	However, most farmers are not capable of accurately identifying pests in the field,
	and there is a limited number of structured data sources available for rapid querying.
	In this work, we explored using domain-agnostic general pre-trained large language model
	(LLM) to extract structured data from agricultural documents with minimal or
	no human intervention.
	We propose a methodology that involves text retrieval and filtering using
	embedding-based retrieval,
	followed by LLM question-answering to automatically extract entities and attributes
	from the documents, and transform them into structured data.
	In comparison to existing methods, our approach achieves consistently better accuracy
	in the benchmark while maintaining efficiency.
\end{abstract}


\section{Introduction}
Information extraction (IE) refers to the process of extracting information from
unstructured text and transform it into structured data.
Nowadays, in an information era, the rapid increase in the amount of data
has made this type of task increasingly important. IE is labour-intensive and
time-consuming, so lots of domains have switched to automatic or semi-automatic
IE \cite{wang_clinical_2018} \cite{saggion_ontology-based_2007}.

The Internet provides a vast amount of information for agriculture,
but the lack of effective data processing methods leads to that
much agricultural information remains unarchived,
buried in news, papers, and government and organization websites.
This may mainly be due to the shortage of annotated corpora \cite{nismi_mol_review_2023}.
These documents cannot be easily analyzed or queried
in their raw form and require some form of information extraction to be easily utilised
in applications. Searching and managing this unstructured information efficiently
is not only a difficult challenge for farmers, but for agriculture professionals as well.

Traditional IE usually require domain-specific training
\cite{drury_survey_2019} \cite{zheng_coreference_2011} or the use of
handcrafted patterns \cite{kaushik_automatic_2018}.
This is challenging for a large number of heterogeneous documents,
coupled with the fact that agricultural professionals may not be
familiar with information extraction.
This leads to effective information being drowned in the flood of information,
which is disappointing.
However, recent research on large language models (LLMs) such as
LLaMA \cite{touvron_llama_2023} and ChatGPT \cite{noauthor_introducing_nodate}
may change this reality.

Large Language Models (LLMs) are deep learning models
that are pre-trained on very large corpora,
capable of understanding natural language, and completing various forms of tasks.
Recent research on LLMs has shown great ability in various NLP areas
\cite{deng_dom-lm_2022} \cite{kojima_large_2023} \cite{wu_bloomberggpt_2023}.
Some papers about IE reveal that their performance bypasses traditional methods
\cite{ma_llm-pruner_2023} \cite{agrawal_large_2022} \cite{arora_language_2023}.

Based on the previous information, this work proposes a system, FINDER.
which is based on general LLM and an embedding-based retrieval (EBR) method,
for zero-shot information extraction from unstructured documents.
The system is capable of automatically extracting entities,
attributes, and corresponding descriptive terms from text
and binding them, generating easily usable structured data.
The system does not require users to know about NLP,
greatly reducing the workload for users.
This makes information extraction tasks related to
agriculture much simpler.

\section{FINDER}

\begin{figure*}[h]
	\centering
	\includegraphics[width=\linewidth]{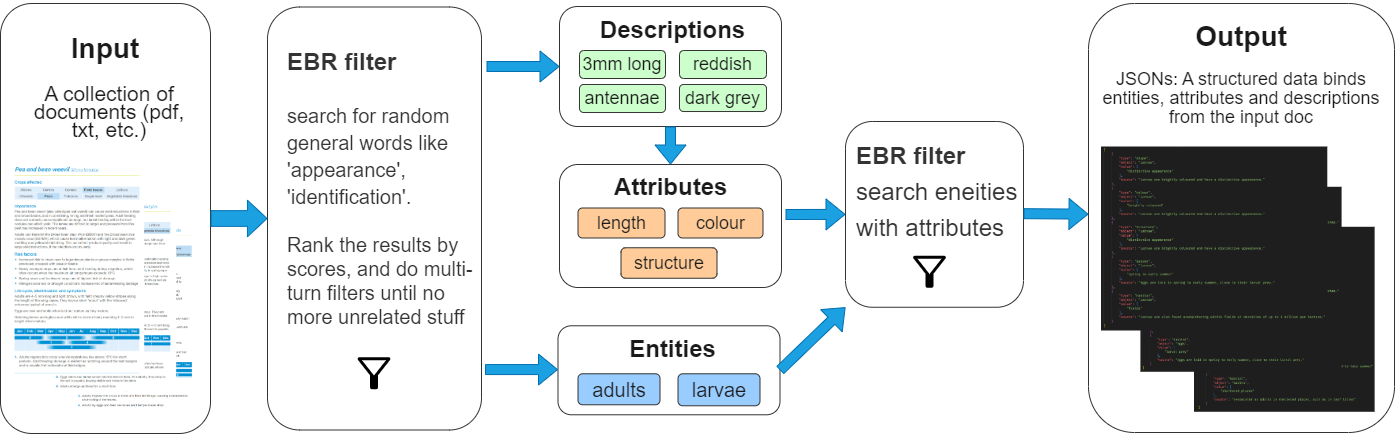}
	\caption{Illustration for the system. The user provides a collection of
		raw documents (e.g. pest data), after 4 steps the system outputs a collection
		of JSON files, which contains all the entities, their attributes and descriptions}
	\label{fig:finder-overview}
\end{figure*}

In this FINDER system, IE tasks are decomposed
into a 4-stage, multi-turn question-and-answer process,
interspersed with EBR to extract relevant text,
avoiding token restrictions and reducing costs.
In the first stage, the system searches for words
used to describe entities in the text.
Then, in the second stage, it identifies all entities that have
been described.
In the third stage, it extracts the attributes represented
by the descriptive words, and finally the fourth stage,
it searches the text for words that describe these attributes
and binds them to the entities, forming structured data.
The overview of the system is shown in \ref{fig:finder-overview}.

\subsection{EBR filter}
The input documents are split into small pieces, vectorised and stored in a vector database.
Text segmentation is necessary here because there is a character limitation for
the LLM input. This still can be a cost-effective approach even if the document is short enough
to be fed directly.

Embedding-based retrieval (EBR) differs from plain text search
in that, it transforms documents into vectors and maps
them to a vector space. This allows similar documents
to have a closer distance in the vector space,
while dissimilar documents have a greater distance.
This enables similarity to be determined by calculating the
distance between documents and search content,
thus completing the retrieval task.

To be prepared for search content with relativeness,
the vector representations of documents are generated and stored in a vector database.
Then, query the database with some general words, for instance: 'appearance, identification'.
This query should perform a k-nearest neighbour (\(kNN\)) search and return multiple
most likely results with their distances to the query words. However these results
may contain unrelated contents, because there is no such standard distance to filter them.

The results, which are text segments, are ranked by distances, from the smallest to the largest,
and a process similar to binary search is applied.
The list is divided into two sets based on the median of the distances,
with the set containing shorter distances referred to as set A and
the other set referred to as set B.
The result located at the median position is extracted to be
checked manually or through LLM to determine whether it contains
the relevant information described by the query.
If the result at the median position satisfies the condition,
then all results in set A are considered to satisfy the condition, and vice versa.
\begin{itemize}
	\item {median point satisfie}: the process is repeated on set B until satisfaction is achieved.
	\item {median point not satisfie}: the process is repeated on set A until satisfaction is achieved.
\end{itemize}
The level of satisfaction is dependent on the user
and will only affect the number of output results.
From the experiments,
the number of results here doesn't affect the final output much,
repeating the process by 2 or 3 times is considered good enough.

\subsection{Stage 1}
The objective of this stage is to use LLM to
identify all descriptive words in the given text.
The filtered text returned from EBR is the input of this step,
a single-turn question and answer is performed for each text segment.
The text is used as context in the prompt template to generate a prompt for LLM.
The question presented to LLM requires it to read
the context and identify all words used to describe entities
and return matched elements as a list.
If no elements meet the criteria, an empty list is returned.
Here 2 example sentences are selected for example, from the chapter "Cabbage stem flea beetle"
in the AHDB database \cite{noauthor_home_nodate}.

\begin{itemize}
	\item \textbf{Example 1:} \textit{
		      Adults are 3–5 mm long, metallic blue-black or light brown.
		      They have long antennae, large hind legs and jump when disturbed.
	      }
	\item \textbf{Example 2:} \textit{
		      Larvae are white, with very small dark spots on the back, a black head and tail
		      and three pairs of dark legs. They can reach 6 mm in length when mature
	      }
\end{itemize}

\subsection{Stage 2}
The objective of this stage is to convert the descriptive words collected from
the stage 1 into attributes. The prompt is generated by passing the
descriptive words as context, followed by a question asking LLM to find
the types of attributes these words represent.
Some examples are given in Table \ref{tab:word-attr}.
The conversion table is observed from the output of LLM,
and may not be consistent across all text depends on the prompts used.
From the above 2 examples, these table is extracted.

\begin{table}[ht]
	\caption{Sample descriptive words to attributes convertion}
	\label{tab:word-attr}
	\centering
	\begin{tabular}{|c|p{0.45\textwidth}|}
		\hline
		\textbf{Descriptive word} & \textbf{Attributes} \\
		\hline
		metallic blue-black, light brown,               \\ white, small dark spots & colour                                                                        \\
		\hline
		6 mm in length            & size                \\
		\hline
		long antennae, large hind legs,                 \\ black head and tail, three pairs of dark legs         & body part\\
		\hline
		jump when disturbed       & behavior            \\
		\hline
		long antennae             & antennae            \\
		\hline
		large hind legs           & legs                \\
		\hline
	\end{tabular}
\end{table}

The inconsistency of the conversion process introduces a problem:
LLM may provide similar but not identical attributes
for similar descriptive words.
For instance, the words "large hind legs," LLM have a high probability
of identifying the attribute as "body part,"
but there is also a small probability of identifying it as
a standalone "legs", and the "antennae" shows the same problem.
While multiple rounds of questioning can be used to extract
the most frequently occurring attribute for a given context,
this approach can easily increase the cost of this step by multiple times,
which is not desirable.

To address this issue, one approach is to aggregate all of the
identified attributes and use LLM to filter them,
retaining the most frequently occurring attribute among
those with similar meanings.
This method can help avoid redundancy while minimising cost increases.

\subsection{Stage 3}

In this step, the utilization of LLM is required to extract the subject from the text,
which corresponds to the Named Entity Recognition (NER) technique in Natural Language Processing (NLP).
Generally, in NLP, the term "entity" encompasses all referable entities,
extending beyond objects without physical shapes,
such as time and race.
However, when processing agricultural and insect-related information with FINDER,
only physical objects are paid attention, particularly those with descriptions.
In the following example text, though lots of entities exist,
the only entities extracted are \textit{Adults} and \textit{Larvae},
the others \textit{antennae, legs, head} are intended to be ignored by the LLM as they are
not the main subject in the paragraph.

\subsection{Stage 4}

In this stage, the EBR filter is once again utilised.
Based on the results obtained from the preceding steps,
possible attributes and a list of entities that have been described within the raw data are all identified.
The subsequent task involves submitting all sentences containing the entity under investigation,
as well as the attributes we aim to identify, to the LLM.
The role of the LLM is to match the entities and attributes,
and return a document in JSON format, the result is shown in Table \ref{tab:output}.

\begin{table}[h]
	\centering
	\begin{tabular}{|c|p{0.8\textwidth}|}
		\hline
		\textbf{Type} & \textbf{Value}                                                                                                                                                                                                     \\
		\hline
		Colour        & Metallic blue-black, light brown                                                                                                                                                                                   \\
		\hline
		Size          & 3–5 mm                                                                                                                                                                                                             \\
		\hline
		Behavior      & Emerge and feed on foliage, rest in moist, sheltered places, migrate into crops to mate and feed on foliage, causing characteristic shot-holing of the leaves, lay eggs and feed on leaves until temperatures drop \\
		\hline
		Antennae      & Long                                                                                                                                                                                                               \\
		\hline
		Legs          & Large hind legs                                                                                                                                                                                                    \\
		\hline
	\end{tabular}
	\caption{The output JSON}
	\label{tab:output}
\end{table}

\section{Experiments}

The system's pipeline consists of six parts, two of which are the EBR filter algorithm,
and the last step is results output.
The attribute extraction (stage 2), entity extraction (stage 3), and the final attribute-entity
matching (stage 4) are completed using the LLM.
Experiments will separately evaluate the performance of these three parts.

Due to the characteristics of the EBR filter,
it only returns results that meet the requirements, and there is no significance for evaluation,
so it is not taken into consideration.

In each test, multiple texts from AHDB\cite{noauthor_home_nodate} will be subjected to
the LLM \textit{gpt-3.5-turbo} with LongChain framework\cite{chase_langchain_2022},
and human evaluators will assess the output of the LLM as "true", "false".
Therefore, the precision rate, recall rate, and F1 score can be calculated.
In the stage 2 and stage 4, another standard "acceptable" is introduces in addition,
the term "acceptable" implies that although there might be discrepancies
between the LLM output and the manually processed text,
the answer can still be considered correct, and will be added to ture positive. For instance,
\textit{adults in sheltered places}, the words \textit{sheltered places}
can be classified as \textbf{location} or \textbf{habitat}, both are correct,
if the manually processed text treat one as correct,
the another one will be an acceptable answer automaticlly.
In real world problem, this may occur with more than 2 seemly correct answer,
and there will be multiple acceptable answer.

\subsection{Results}

\begin{table}[b]
	\centering
	\begin{tabular}{|l|l|l|l|}
		\hline
		                       & \textbf{Stage 2} & \textbf{Stage 3} & \textbf{Stage 4} \\ \hline
		\textbf{Precision}     & 76\%             & 89\%             & 88\%             \\ \hline
		\textbf{Precision AC.} & 90\%             & -                & 90\%             \\ \hline
		\textbf{Recall}        & 72\%             & 84\%             & 84\%             \\ \hline
		\textbf{Recall AC.}    & 76\%             & -                & 89\%             \\ \hline
	\end{tabular}
	\caption{\textbf{Precision and Recall from Stage 2,3,4 at zero-shot}. The precision AC. and Recall AC. are precision/recall with acceptable answers.
		The answers are checked by human and compared with human annotated answers}
	\label{tab:results}
\end{table}

Table \ref{tab:results} shows the performance of gpt3.5-turbo under zero-shot conditions.
The model demonstrates a high accuracy in stage 3,
namely entity extraction, but does not perform as well in the other two aspects.
However, after deeming some answers as "acceptable", the accuracy have significant increases,
with a maximum improvement of \(13\%\).

This is an interesting phenomenon. In fact, many things can be interpreted in multiple ways,
and there is no need to adhere strictly to a single correct answer.
However, the standard for what is "acceptable" varies from person to person,
so this experiment does not fully demonstrate that the LLM has a
very high accuracy when outputting acceptable answers.

\section{Conclution}

In this work, we explored the use of LLM to extract information from unlabeled agricultural datasets.
The results indicate that LLM has great potential in the task of agricultural information extraction.
It can extract data into easily accessible JSON documents with high accuracy in zero-shot tasks,
requiring little or no human intervention.

A pipeline workflow for using LLM to perform information extraction tasks is introduced in this work.
By breaking down a large task into several smaller ones in the form of a pipeline,
it become capable of doing performance evaluation and optimisation for each small individual task.

While this project already has some capability for information extraction,
whether it can be practically applied remains an unknown, and there is still a lot of room for improvement.
In testing, we found that the output from the LLM does not always full-fill the requirement,
as demonstrated by the following points:

\begin{itemize}
	\item When converting descriptive text into attributes,
	      the output attributes often seem somewhat off.
	      For example, "dark spots" could be identified as "colour" or "pattern".
	      This distinction can sometimes be considered acceptable,
	      although this standard of acceptability is very subjective,
	      causing the accuracy of the entire pipeline to fluctuate.

	\item The LLM's output is also not entirely satisfactory when matching values and attributes.
	      For example, when the value for "size" is "6mm in length when mature",
	      the ideal output should be just "6mm".
\end{itemize}

These minor errors could potentially cause some problems,
but without large-scale testing or practical application,
this is difficult to ascertain.
Moreover, this issue doesn't seem to be well circumvented by modifying prompts,
and it might require a model specifically trained for information extraction
rather than a general chat model to solve it.

The future work of this project will focus on how to adjust the prompts and the structure
of the pipeline for more precise information extraction,
trying to resolve the above issues through research in these two areas.
It will also require the use of more LLMs and datasets for testing to determine
the stability of this pipeline.

\bibliographystyle{unsrtnat}
\bibliography{references}  






\end{document}